\documentclass[11pt]{article}
\usepackage[preprint]{acl}

\usepackage{times}
\usepackage{latexsym}
\usepackage[T1]{fontenc}
\usepackage[utf8]{inputenc}
\usepackage{microtype}
\usepackage{inconsolata}
\usepackage{graphicx}
\usepackage{booktabs}

\usepackage{amsmath}
\usepackage{amssymb}
\usepackage{xcolor}
\usepackage{fvextra}

\title{Is Jev a Good Multilingual Judge?}

\author{
  Wei Liu\textsuperscript{1}, Yi Fan\textsuperscript{1}, Xiyan Fu\textsuperscript{2}\thanks{Corresponding author.} \\
  \textsuperscript{1}Heidelberg Institute for Theoretical Studies \\
  \textsuperscript{2}Independent Researcher \\
  \texttt{wei.liu.llm@gmail.com}
}

\begin{document}
\maketitle

\begin{abstract}
Generative large language models are widely used for multilingual text evaluation, yet whether decision models that directly output judgment probabilities can serve as multilingual judges has received limited attention. We evaluate Jev against generative LLM judges on four multilingual benchmarks with human annotations, covering up to 47 language varieties and three evaluation formats: binary decisions, pairwise comparisons, and graded ratings. At approximately $1/31$ to $1/61$ of GPT-6.1's evaluation cost, Jev is competitive on most multilingual judging tasks, but falls behind GPT-6.1 on knowledge-intensive answer verification. Further analysis shows that Jev is more likely to accept incorrect answers in low-resource languages than in English and that its probability outputs help identify errors across tasks. Based on these findings, we offer recommendations on evaluation instructions, confidence-based selection, and model routing, providing empirical guidance for using Jev as a multilingual judge.
\end{abstract}

\section{Introduction}
\label{sec:intro}

Generative large language models are widely used as judges to evaluate model outputs, drawing on their strong language understanding and reasoning capabilities \citep{zheng2023judging}. Many evaluation tasks, however, require only binary decisions, pairwise preferences, or graded ratings, rather than free-form text. For these tasks, decision models such as Jev \citep{typesafe2026jev} offer an alternative by directly predicting probabilities over possible judgments. Evaluations on English benchmarks have shown that Jev can achieve performance comparable to generative judges on some evaluation tasks at substantially lower cost, highlighting its potential as an automated judge \citep{li2026jevjudge,rao2026rubric}.

As large language models are increasingly used across languages, automated judges must also support multilingual evaluation. However, existing evaluations of Jev as a judge have focused primarily on English \citep{li2026jevjudge,rao2026rubric}, while multilingual evaluations have examined its reading comprehension and question-answering abilities \citep{deusser2026jev,li2026specialized}. Its ability to evaluate model outputs across languages has yet to be systematically studied. It therefore remains unclear how reliably Jev can serve as a multilingual judge across languages and evaluation tasks.

To fill this gap, we systematically evaluate Jev against generative LLM judges on four multilingual benchmarks: MEMERAG for RAG faithfulness \citep{blandon2025memerag}, Global-MMLU-Lite for answer verification \citep{singh2025globalmmlu}, MENLO for response quality \citep{whitehouse2026menlo}, and METAL for summary quality \citep{hada-etal-2024-metal}. These benchmarks cover up to 47 language varieties and three evaluation formats: binary decisions, pairwise comparisons, and graded ratings. We assess judgment quality against human annotations and use parallel data to examine whether judgments remain consistent across languages.

Our results show that Jev is competitive on most multilingual judging tasks at approximately $1/31$ to $1/61$ of GPT-6.1's evaluation cost, but falls behind GPT-6.1 on knowledge-intensive answer verification. On parallel data, Jev's judgments are more consistent across languages than Qwen's, but less consistent than GPT-6.1's, particularly in low-resource languages. Further analysis shows that the gap in low-resource languages is primarily driven by Jev accepting incorrect answers that it correctly rejects in English.

Beyond judgment quality, we examine whether Jev's probability outputs can help identify potentially incorrect judgments. We find that the confidence assigned to a judgment helps distinguish correct from incorrect decisions, although this distinction is clearer in binary verification than in graded ratings. We further investigate how these findings can inform the practical use of Jev. Our results support 1) using English evaluation instructions by default; 2) retaining only high-confidence judgments when accuracy takes priority over coverage; and 3) selectively routing low-confidence samples to models that perform better on the target task for additional performance gains, although these gains are modest on most tasks.

Our contributions are threefold:
\begin{itemize}
    \item We systematically compare Jev with generative LLM judges on four multilingual benchmarks covering up to 47 language varieties and three evaluation formats.
    \item We analyze the judgment changes underlying cross-lingual inconsistency and assess whether Jev's decision probabilities help distinguish correct from incorrect judgments.
    \item We provide practical recommendations for using Jev through experiments on instruction language, confidence-based selection, and model routing.
\end{itemize}

\section{Related Work}
\label{sec:related}

\paragraph{Multilingual LLM Judges.}
Generative LLMs are widely used as automated judges \citep{zheng2023judging}, with evaluation increasingly extending to multilingual settings. Benchmarks such as METAL \citep{hada-etal-2024-metal}, MM-Eval \citep{son2025mmeval}, M-RewardBench \citep{gureja2025mrewardbench}, and MENLO \citep{whitehouse2026menlo} assess agreement between model judgments and human annotations across tasks and languages. Prior studies report variation in judge performance across languages \citep{son2025mmeval}, more lenient ratings in lower-resource languages \citep{zhou2026lower}, and inconsistent judgments of the same content across languages \citep{fu2025multilingual}. Pairwise judgments are also sensitive to candidate presentation order \citep{zheng2023judging,wang2023fair}. Other work explores dedicated training to improve multilingual judging \citep{pombal2025mprometheus,doddapaneni2025cross}. We examine the capabilities and behavior of decision models that directly output judgment probabilities in multilingual evaluation.

\paragraph{Decision Models as Judges.}
Existing evaluations of Jev \citep{typesafe2026jev} fall into two broad categories. The first examines its general decision-making capabilities. \citet{deusser2026jev} evaluate Jev across classification, moderation, rubric scoring, and other tasks, while \citet{li2026specialized} examine its knowledge, reasoning, and multilingual understanding. The second investigates Jev's ability to evaluate other models' outputs. \citet{li2026jevjudge} study judgment quality and confidence-based escalation, while \citet{rao2026rubric} compare decision models with generative judges under shared evaluation criteria. These studies report competitive judging performance at lower cost in several settings, but focus primarily on English, leaving Jev's multilingual applicability unclear. We extend this line of work by evaluating Jev across languages and judgment formats, examining both judgment quality and cross-lingual consistency.

\paragraph{Probability Reliability and Judge Cascades.}
The reliability of probability estimates concerns both their correspondence with observed correctness and their ability to identify likely errors. Prior work examines calibration differences across languages \citep{ahuja2022calibration}, while research on model cascades uses confidence to refer uncertain judgments to other models, balancing quality and cost \citep{chen2023frugalgpt,jung2024trust}. Studies of Jev find this strategy effective in some English evaluations \citep{li2026jevjudge}, but report limited gains when judges share many of the same errors \citep{rao2026rubric}. We examine the reliability of Jev's probability outputs in multilingual evaluation and evaluate confidence-based routing to other judges.

\begin{table*}[t]
\centering
\small
\setlength{\tabcolsep}{7pt}
\begin{tabular}{@{}lccccc@{}}
\toprule
\raisebox{-7pt}{Benchmark} & \raisebox{-7pt}{Task} & \raisebox{-7pt}{Languages} & \multicolumn{3}{c}{Evaluation format} \\
\cmidrule(lr){4-6}
& & & Binary & Pairwise & Graded \\
\midrule
MEMERAG & RAG faithfulness & 5 & $\checkmark$ & $\times$ & $\times$ \\
Global-MMLU-Lite & Answer verification & 23 & $\checkmark$ & $\times$ & $\times$ \\
MENLO & Response quality & 47 & $\times$ & $\checkmark$ & $\checkmark$ \\
METAL & Summary evaluation & 10 & $\checkmark$ & $\times$ & $\checkmark$ \\
\bottomrule
\end{tabular}
\caption{Overview of the four multilingual benchmarks.}
\label{tab:datasets}
\end{table*}

\section{Experimental Setup}
\label{sec:setup}

\subsection{Benchmarks}
\label{sec:data}

We evaluate Jev on four multilingual benchmarks covering answer verification, RAG faithfulness, response quality, and summary quality. These benchmarks support three evaluation formats---binary decisions, pairwise comparisons, and graded ratings---as summarized in Table~\ref{tab:datasets}.

\paragraph{MEMERAG.}
We evaluate the faithfulness of model responses to retrieved passages at the sentence level \citep{blandon2025memerag}. For each judgment, the judge receives a sentence from the response, together with the full response and retrieved passages, and determines whether that sentence is grounded in the retrieved evidence and relevant to the question. We evaluate these judgments against reference labels for 2,034 sentences across five languages.

\paragraph{Global-MMLU-Lite.}
We use multilingual multiple-choice questions \citep{singh2025globalmmlu} to evaluate whether judges can distinguish correct from incorrect answers. Specifically, we construct two verification samples from each question by pairing it with either the correct answer or a randomly selected distractor. The judge sees the full question, including all four options, and determines whether the paired answer is correct. We use the same correct and distractor options across the dataset's language versions, allowing us to examine whether judgments remain consistent across languages.

\paragraph{MENLO.}
We use MENLO's human ratings of response quality across 47 language varieties \citep{whitehouse2026menlo} to construct two evaluation tasks: graded rating and pairwise comparison. Specifically, the graded-rating task asks the judge to assign a response one of five levels according to the dataset's evaluation criteria, using the human rating as the reference. For pairwise comparison, the judge selects the better of two responses to the same prompt. To construct these pairs, we exclude pairs with equal human ratings and treat the higher-rated response in each pair as preferred.

\paragraph{METAL.}
We evaluate summary quality in ten languages through hallucination detection and graded rating \citep{hada-etal-2024-metal}. Specifically, the judge receives a summary and its source article and determines whether the summary contains hallucinations. It also assigns separate three-level ratings for linguistic acceptability, output content quality, and task quality. We evaluate these judgments against reference labels aggregated from three human annotators.

Further preprocessing and sample-selection details are provided in Appendix~\ref{app:data}.

\subsection{Judges and Evaluation Protocol}
\label{sec:judges}
\label{sec:protocol}

We compare Jev (\texttt{jev-1.13.0}) with two generative LLM judges, Qwen3.8-27B (Qwen) and GPT-6.1. All judges receive the same task instructions, evaluation criteria, and content. Instructions are written in English unless otherwise specified, while the evaluated content remains in its original language.

Jev directly outputs decision probabilities, whereas the generative judges are instructed to return only the requested judgment without an explanation. We disable thinking for Qwen and use low reasoning effort for GPT-6.1. Although both generative judges return discrete judgments, we additionally derive answer probabilities from Qwen's output-token probabilities.

For binary verification and graded rating, GPT-6.1's judgments are taken directly from its discrete outputs. For Jev and Qwen, we obtain binary judgments using a default acceptance threshold of 0.5 and graded ratings by selecting the most probable level. For pairwise comparison, we evaluate each response pair in both presentation orders. We align Jev's and Qwen's predictions to the same candidate and average their probabilities to determine the final preference. For GPT-6.1, matching choices across the two orders yield a preference, while conflicting choices are treated as a tie.

Prompt templates are provided in Appendix~\ref{app:prompts}, and further inference and output-processing details are given in Appendix~\ref{app:inference}.

\begin{table*}[!t]
\centering
\small
\setlength{\tabcolsep}{9pt}
\begin{tabular}{@{}lcccccc@{}}
\toprule
\raisebox{-7pt}{Judge} & \multicolumn{2}{c}{MEMERAG} & \multicolumn{2}{c}{Global-MMLU-Lite} & \multicolumn{2}{c}{METAL (Hallucination)} \\
\cmidrule(lr){2-3}\cmidrule(lr){4-5}\cmidrule(lr){6-7}
& BAcc $\uparrow$ & AUROC $\uparrow$ & BAcc $\uparrow$ & AUROC $\uparrow$ & BAcc $\uparrow$ & AUROC $\uparrow$ \\
\midrule
Jev & 69.0 & 81.0 & 89.6 & 95.8 & \textbf{85.9} & \textbf{92.6} \\
Qwen & \textbf{74.1} & \textbf{82.0} & 82.9 & 91.0 & 84.9 & 92.3 \\
GPT-6.1 & 73.5 & --- & \textbf{96.1} & --- & 79.2 & --- \\
\bottomrule
\end{tabular}

\addtocounter{table}{1}
\caption{Performance on RAG faithfulness, answer verification, and hallucination detection. Dashes indicate unavailable answer probabilities for GPT-6.1.}
\label{tab:main}

\end{table*}

\begin{table}[t]

\centering
\small
\setlength{\tabcolsep}{4pt}
\resizebox{\columnwidth}{!}{\begin{tabular}{@{}lccc@{}}
\toprule
Benchmark & Jev & Qwen & GPT-6.1 \\
\midrule
MEMERAG & \textbf{0.082} & 1.240 (15.2$\times$) & 5.010 (61$\times$) \\
Global-MMLU-Lite & \textbf{0.026} & 0.160 (6.2$\times$) & 0.820 (32$\times$) \\
MENLO & \textbf{0.081} & 0.620 (7.8$\times$) & 2.950 (37$\times$) \\
METAL & \textbf{0.063} & 0.440 (7.1$\times$) & 1.960 (31$\times$) \\
\bottomrule
\end{tabular}
}
\addtocounter{table}{-2}
\caption{Evaluation cost per 1,000 samples (USD). Parentheses indicate how many times each model costs as much as Jev.}
\label{tab:efficiency}
\addtocounter{table}{1}

\end{table}

\subsection{Evaluation Metrics}
\label{sec:metrics}

We evaluate judges in terms of judgment quality and consistency. Judgment quality measures agreement with human annotations or reference answers. For binary tasks, we report balanced accuracy (BAcc), defined as the mean of positive- and negative-class recall, and AUROC where answer probabilities are available. For pairwise comparisons, we report preference accuracy, assigning 1 to a correct preference, 0 to an incorrect preference, and 0.5 to a tied prediction. For graded ratings, we report macro-F1 and Spearman correlation with human ratings.

Judgment consistency measures whether judgments remain unchanged when the same content is presented in different languages or orders. For cross-lingual consistency, we report agreement with English on Global-MMLU-Lite, defined as the proportion of candidate answers receiving the same judgment in English and a target language. We also report Fleiss' $\kappa$ to measure agreement across all 23 language versions. For order consistency, we evaluate each pair of candidate responses in MENLO in both presentation orders, A--B and B--A, and calculate the proportion of pairs for which the judge selects the same response in both orders.

Details of metric aggregation are provided in Appendix~\ref{app:stats}.

\section{Multilingual Judging Performance}
\label{sec:capability}

\subsection{Evaluation Cost}
\label{sec:evaluation-cost}

We compare the three judges' evaluation costs on the same samples within each benchmark, calculated from recorded token usage and API prices. Table~\ref{tab:efficiency} reports costs per 1,000 samples. Jev costs \$0.026--\$0.082, compared with \$0.160--\$1.240 for Qwen and \$0.820--\$5.010 for GPT-6.1. Across all 43,484 samples, the total costs are \$2.64, \$20.96, and \$97.45, respectively. Qwen and GPT-6.1 therefore cost 7.9 and 36.9 times as much as Jev over the full evaluation. See Appendix~\ref{app:noise-cost} for cost-calculation details.

\begin{table}[t]

\centering
\small
\begin{tabular}{@{}lc@{}}
\toprule
Judge & Preference accuracy (\%) $\uparrow$ \\
\midrule
Jev & 74.6 \\
Qwen & \textbf{74.7} \\
GPT-6.1 & 72.0 \\
\bottomrule
\end{tabular}

\caption{Pairwise preference accuracy on MENLO, using preferences derived from human ratings.}
\label{tab:pairwise}

\end{table}

\subsection{Judgment Quality}
\label{sec:judgment-quality}

\begin{table*}[t]
\centering
\small
\setlength{\tabcolsep}{11pt}
\begin{tabular}{@{}lcccc@{}}
\toprule
\raisebox{-7pt}{Judge} & \multicolumn{2}{c}{MENLO (Response rating)} & \multicolumn{2}{c}{METAL (Summary rating)} \\
\cmidrule(lr){2-3}\cmidrule(lr){4-5}
& Macro-F1 $\uparrow$ & Spearman $\rho\uparrow$ & Macro-F1 $\uparrow$ & Spearman $\rho\uparrow$ \\
\midrule
Jev & \textbf{33.3} & 0.29 & 55.0 & 0.60 \\
Qwen & 31.9 & \textbf{0.31} & \textbf{56.2} & \textbf{0.65} \\
GPT-6.1 & 28.3 & 0.30 & 55.5 & 0.63 \\
\bottomrule
\end{tabular}

\caption{Response rating on MENLO and summary rating on METAL. Macro-F1 measures agreement with reference rating levels, while Spearman's $\rho$ measures rating correlation.}
\label{tab:graded}
\end{table*}

%
%

We next examine Jev's judgment quality. Tables~\ref{tab:main}--\ref{tab:graded} compare its performance with the generative judges across binary decisions, pairwise preferences, and graded ratings. At approximately $1/31$--$1/61$ of GPT-6.1's evaluation cost, Jev performs competitively on several tasks, achieving 74.6\% accuracy on MENLO preference judgments, close to Qwen and 2.6 percentage points above GPT-6.1. It also achieves the highest balanced accuracy on METAL hallucination detection at 85.9\% and the highest macro-F1 on MENLO ratings at 33.3\%.

However, this competitiveness does not extend uniformly across tasks or metrics. Jev trails both generative judges in balanced accuracy on MEMERAG and falls 6.5 percentage points below GPT-6.1 on Global-MMLU-Lite answer verification. Moreover, Jev's higher macro-F1 on MENLO does not translate into stronger rating correlation. Its Spearman correlation is similar to the baselines on MENLO and lower on METAL. These results indicate that Jev's competitive performance depends on the evaluation task and does not consistently extend across rating metrics.

\subsection{Judgment Consistency}
\label{sec:judgment-consistency}

\begin{figure}[t]
\centering
\includegraphics[width=\columnwidth]{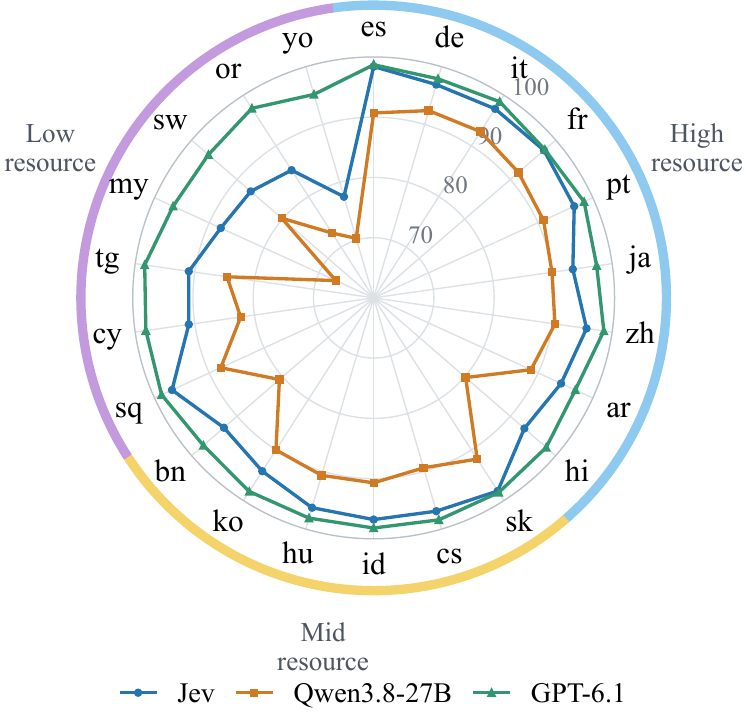}
\caption{Agreement with English judgments across 22 non-English languages in Global-MMLU-Lite. Outer bands indicate high-, mid-, and low-resource language groups defined in the dataset \citep{singh2025globalmmlu}.}
\label{fig:cross-lingual-consistency}
\end{figure}

\begin{table}[t]
\centering
\small
\setlength{\tabcolsep}{3pt}
\resizebox{\columnwidth}{!}{\begin{tabular}{@{}lccc@{}}
\toprule
& \multicolumn{2}{c}{Cross-lingual} & Order \\
\cmidrule(lr){2-3}\cmidrule(lr){4-4}
Judge & Agreement (\%) & Fleiss' $\kappa$ & Agreement (\%) \\
\midrule
Jev & 93.4 & 0.81 & 93.2 \\
Qwen & 85.7 & 0.63 & 76.6 \\
GPT-6.1 & \textbf{97.8} & \textbf{0.95} & \textbf{94.6} \\
\bottomrule
\end{tabular}
}
\caption{Cross-lingual and order consistency. On Global-MMLU-Lite, agreement compares each judge's decisions with its English decisions, while Fleiss' $\kappa$ measures agreement across all language versions. On MENLO, order agreement measures how often a judge selects the same response when the two candidate responses are presented in reverse order.}
\label{tab:consistency}
\end{table}

\paragraph{Cross-lingual consistency.}
We compare judges' consistency across languages using parallel data from Global-MMLU-Lite, measuring agreement as the proportion of candidate answers receiving the same judgment in English and a target language. As shown in Table~\ref{tab:consistency}, Jev achieves an average agreement of 93.4\% across all 22 non-English languages, higher than Qwen's 85.7\% but lower than GPT-6.1's 97.8\%. We further break down agreement by language and present the results in Figure~\ref{fig:cross-lingual-consistency}. Jev maintains high agreement in high- and mid-resource languages, with relatively small gaps to GPT-6.1. However, in most low-resource languages, Jev shows lower agreement and a larger gap to GPT-6.1, although it still outperforms Qwen. For example, Jev's agreement with English falls to 77.5\% in Yoruba. This decline is not uniform across low-resource languages, as Albanian maintains an agreement rate of 96.8\%.

In addition to agreement with English, we use Fleiss' $\kappa$ to compare the three judges' overall consistency across all 23 language versions. Table~\ref{tab:consistency} shows that Jev achieves a higher $\kappa$ value than Qwen but a lower value than GPT-6.1, consistent with the agreement results.

\paragraph{Order consistency.}
We compare judges' consistency on MENLO after swapping the positions of the candidate responses, reporting the proportion of pairs for which the same response is selected in both orders (agreement rate). As shown in Table~\ref{tab:consistency}, Jev achieves an agreement rate of 93.2\%, close to GPT-6.1 and substantially higher than Qwen. These results indicate that Jev's pairwise preference judgments are relatively insensitive to candidate positions, with stability close to that of GPT-6.1.

Overall, Jev offers competitive multilingual judging performance at substantially lower cost, although its relative performance varies across tasks. While its order consistency is close to GPT-6.1's, the gap in cross-lingual consistency is small in high- and mid-resource languages but larger in most low-resource languages.

\section{Inconsistency Analysis}
\label{sec:attribution}
\label{sec:cross-lingual-errors}

\begin{figure*}[!t]
\centering
\includegraphics[width=\textwidth]{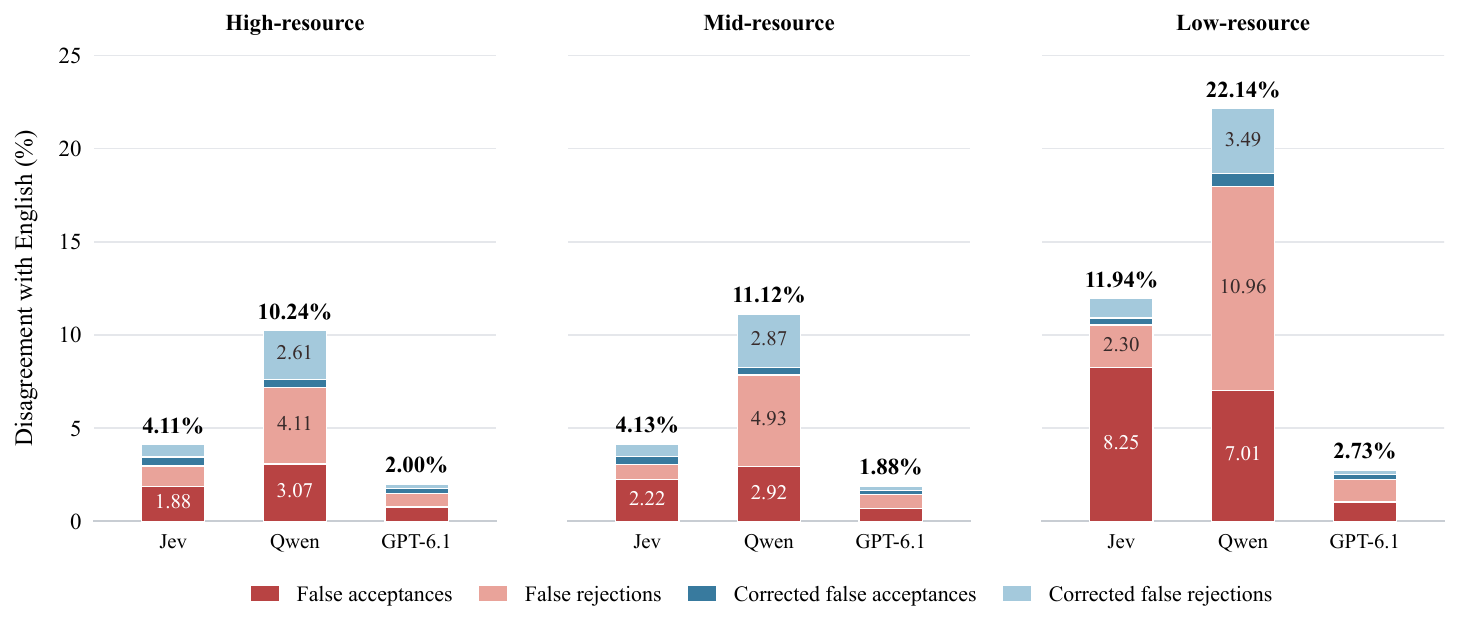}
\caption{Changes in judgments from English to other languages on Global-MMLU-Lite. False acceptances and false rejections are newly introduced errors, while their corrected counterparts are English errors corrected in the target language. Each segment shows a percentage of all candidate answers, averaged across languages in the resource group. Total bar heights give disagreement with English.}
\label{fig:cross-lingual-errors}
\end{figure*}

\begin{figure*}[!t]
\centering
\includegraphics[width=\textwidth]{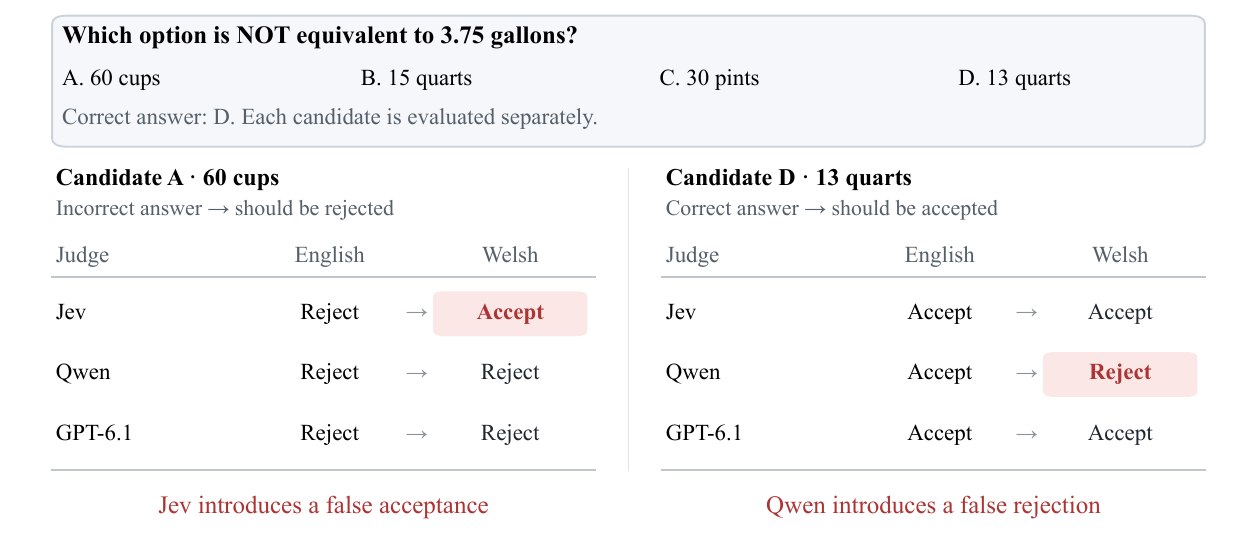}
\caption{A Global-MMLU-Lite example of judgment changes from English to Welsh. Jev introduces a false acceptance and Qwen a false rejection, while GPT-6.1 maintains both correct judgments. Incorrect judgments are highlighted in red.}
\label{fig:unit-conversion-case}
\end{figure*}

Section~\ref{sec:judgment-consistency} shows that Jev's cross-lingual consistency falls between Qwen's and GPT-6.1's, with a larger gap to GPT-6.1 in low-resource languages. However, disagreement with English judgments can reflect either newly introduced errors or corrections of errors made in English. To examine which changes account for the consistency gaps, we analyze all three models' judgments on the full Global-MMLU-Lite evaluation set. Specifically, we decompose judgment changes from English to the target language into four categories: (1) false acceptances, where incorrect answers change from rejected to accepted; (2) false rejections, where correct answers change from accepted to rejected; (3) corrected false acceptances, where incorrect answers change from accepted to rejected; and (4) corrected false rejections, where correct answers change from rejected to accepted. For each language, we divide each category's count by the total number of candidate answers, then average these rates equally across non-English languages within each resource group defined in Section~\ref{sec:judgment-consistency}. Figure~\ref{fig:cross-lingual-errors} shows the results. See Appendix~\ref{app:cross-lingual-transitions} for calculation details.

Most of Jev's disagreement with English comes from newly introduced errors rather than corrections of English errors. This pattern is particularly pronounced in low-resource languages, where 10.5\% of judgments change from correct to incorrect, while only 1.4\% change from incorrect to correct. Among the newly introduced errors, false acceptances account for the larger share. Their contribution rises from 1.88 and 2.22 percentage points in high- and mid-resource languages to 8.25 points in low-resource languages, compared with only 1.03 points for GPT-6.1 in the low-resource group. This difference accounts for about 78\% of the disagreement gap between the two models in that group, indicating that Jev's lower consistency primarily reflects its acceptance of incorrect answers that it correctly rejects in English. By contrast, Qwen's disagreement with English is mainly driven by newly introduced false rejections rather than false acceptances across all three resource groups.

Figure~\ref{fig:unit-conversion-case} illustrates these contrasting error patterns on a Global-MMLU-Lite question. All three models correctly reject candidate A and accept candidate D in English. In Welsh, however, Jev incorrectly accepts A and Qwen incorrectly rejects D, while GPT-6.1 correctly rejects A and accepts D.

\section{Confidence-Based Error Detection}
\label{sec:probabilities}

\begin{figure}[!t]
\centering
\includegraphics[width=\columnwidth]{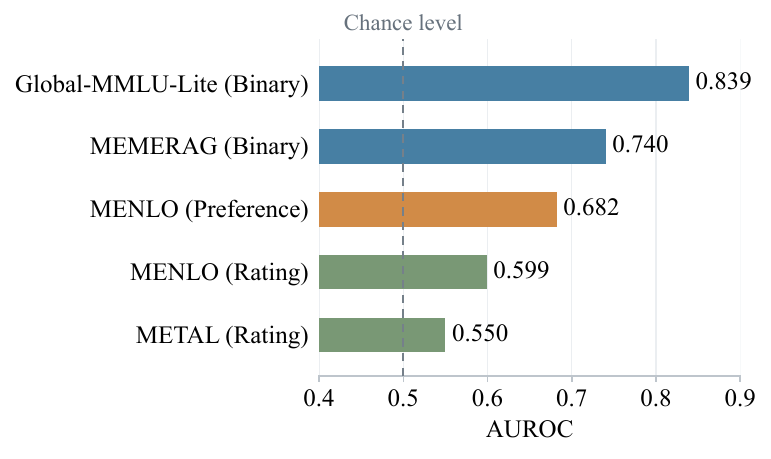}
\caption{How well Jev's confidence distinguishes correct from incorrect judgments, measured by AUROC. Colors distinguish judgment types, and the dashed line marks chance level (0.5).}
\label{fig:confidence-error-analysis}
\end{figure}

Jev outputs not only a judgment but also the probabilities associated with its decisions. We ask whether these probabilities provide useful information about judgment reliability. To examine this, we take the probability assigned to the final judgment as its confidence and test whether correct judgments receive higher confidence than incorrect ones. Specifically, for binary verification and pairwise comparison, we use the probability assigned to the selected decision as confidence, while for graded rating, confidence is defined as the probability assigned to the selected score level. We compute AUROC using judgment correctness as the binary label and confidence as the score, with correct judgments as the positive class. An AUROC of 0.5 indicates chance-level discrimination, while higher values indicate better discrimination. Figure~\ref{fig:confidence-error-analysis} shows the results across three judgment types. See Appendix~\ref{app:confidence-task-details} for more details.

The AUROC results are above 0.5 across all five tasks, indicating that Jev's probability outputs provide additional information for identifying errors, although their usefulness varies across judgment types. For binary verification, AUROC reaches 0.839 on Global-MMLU-Lite and 0.740 on MEMERAG, showing that correct judgments tend to receive higher confidence. Confidence also distinguishes correct from incorrect choices in MENLO pairwise comparison to some extent, with an AUROC of 0.682. By contrast, AUROC is closer to chance level for graded rating, at 0.599 on MENLO and 0.550 on METAL. These results indicate that Jev's probability outputs help identify errors in binary verification, while providing a weaker reliability signal for graded rating.

\section{Practical Guidelines}
\label{sec:usage}

We examine how evaluation instructions, confidence-based selection, and model routing affect Jev's judgment quality and cost, providing practical guidance for its use. See Appendix~\ref{app:guidance-details} for experimental details and supplementary results.

\subsection{Instruction Language}
\label{sec:rubric-language}
\begin{table}[t]
\centering
\small
\begin{tabular}{@{}lcc@{}}
\toprule
Task & English & Target language\\
\midrule
Global-MMLU-Lite & 89.39 & 89.21\\
MENLO (Preference) & 67.42 & 67.02\\
MENLO (Rating) & 30.80 & 31.32\\
\bottomrule
\end{tabular}

\caption{Performance using English instructions or instructions translated into the language of the evaluated content.}
\label{tab:guidance-configuration}
\end{table}

\begin{table*}[t]
\centering
\small
\setlength{\tabcolsep}{4pt}
\begin{tabular}{@{}lcccccccc@{}}
\toprule
& \multicolumn{2}{c}{No filtering} & \multicolumn{2}{c}{$\geq 0.7$} & \multicolumn{2}{c}{$\geq 0.9$} & \multicolumn{2}{c}{$\geq 0.95$}\\
\cmidrule(lr){2-3}\cmidrule(lr){4-5}\cmidrule(lr){6-7}\cmidrule(l){8-9}
Task & Acc. & Cov. & Acc. & Cov. & Acc. & Cov. & Acc. & Cov.\\
\midrule
Global-MMLU-Lite (Verification) & 89.57 & 100.00 & 94.90 & 81.73 & 98.48 & 46.95 & 99.18 & 28.00\\
MENLO (Preference) & 74.63 & 100.00 & 78.66 & 82.11 & 83.97 & 53.10 & 86.69 & 38.24\\
MENLO (Rating) & 35.91 & 100.00 & 46.89 & 30.49 & 62.86 & 3.81 & 69.27 & 1.72\\
\bottomrule
\end{tabular}

\caption{Accuracy of retained judgments (Acc.) and percentage of samples retained (Cov.) at different confidence thresholds. ``No filtering'' retains all judgments, while $\geq t$ retains only judgments with confidence at least $t$.}
\label{tab:guidance-confidence-thresholds}
\end{table*}


\begin{table*}[t]
\centering
\small


\begin{tabular*}{\textwidth}{@{\extracolsep{\fill}}lccccc@{}}
\toprule
Task & Jev & Jev $\rightarrow$ GPT & Jev $\rightarrow$ GPT & Jev $\rightarrow$ Qwen & Jev $\rightarrow$ Qwen\\
 & & (random) & (confidence) & (random) & (confidence)\\
\midrule
Global-MMLU-Lite & 89.57 & 90.22 & 92.73 & 88.90 & 89.90\\
MENLO (Pairwise comparison) & 74.64 & 74.51 & 74.64 & 74.44 & 75.04\\
METAL (Summary rating) & 55.01 & 55.12 & 55.67 & 55.28 & 55.69\\
\bottomrule
\end{tabular*}

\caption{Performance of Jev alone and routing to GPT or Qwen. ``Confidence'' selects test samples below a threshold determined by the lowest-confidence 10\% of development samples, while ``random'' selects the same number of test samples at random.}
\label{tab:guidance-routing}
\end{table*}

Multilingual judging requires evaluating content in different languages, but it is unclear whether evaluation instructions also need to match the language of that content. We compare English and target-language instructions across 22 non-English languages in Global-MMLU-Lite and 8 languages in MENLO, keeping the evaluated content and evaluation standards unchanged. As shown in Table~\ref{tab:guidance-configuration}, translating the instructions changes BAcc from 89.39\% to 89.21\% for answer verification and preference accuracy from 67.42\% to 67.02\% for pairwise comparison. For graded rating, macro-F1 increases slightly, but correlation with human ratings does not improve (see Appendix~\ref{app:native}). Overall, translation provides no clear performance gains, while increasing mean input token counts by 13.5\%, 27.1\%, and 38.6\% for the three judgment types, respectively. We therefore recommend using English evaluation instructions by default for multilingual judging, without translating them into every target language.

\subsection{Confidence-Based Selection}
\label{sec:confidence-selection}

Section~\ref{sec:probabilities} shows that Jev's confidence helps distinguish correct from incorrect judgments. To assess how this signal can be used when higher accuracy is required, we compare the accuracy of judgments retained at different confidence thresholds on Global-MMLU-Lite answer verification, MENLO pairwise comparison, and MENLO response rating. Table~\ref{tab:guidance-confidence-thresholds} shows that higher thresholds improve accuracy on all three tasks, but both the gains and the proportion of samples retained vary across tasks. Raising the threshold from 0.7 to 0.95 improves accuracy from 94.90\% to 99.18\% for answer verification and from 78.66\% to 86.69\% for pairwise comparison, while still retaining 28.00\% and 38.24\% of samples, respectively. In contrast, response rating achieves a larger accuracy gain, from 46.89\% to 69.27\%, but retains only 1.72\% of samples. We therefore recommend raising the confidence threshold when accuracy takes priority over coverage. When both matter, users should choose a threshold that maximizes coverage while meeting their target accuracy, particularly for rating tasks.

\subsection{Model Routing}
\label{sec:routing}

Section~\ref{sec:probabilities} shows that Jev's confidence helps distinguish correct from incorrect judgments. We therefore examine whether routing low-confidence samples to another model improves overall performance. Given GPT's stronger overall performance in Section~\ref{sec:capability}, we primarily consider routing to GPT and include Qwen for comparison. To determine which samples to route, we select a confidence threshold on the development set such that approximately 10\% of samples fall below it. We then apply this threshold to the test set, using the selected model's judgments for samples below the threshold and retaining Jev's judgments for the rest. For comparison, we randomly route the same number of samples within each test fold.

Table~\ref{tab:guidance-routing} shows that routing low-confidence samples improves performance to different degrees across the three tasks. On Global-MMLU-Lite, routing to GPT increases performance from 89.57\% to 92.73\%, while routing to Qwen yields a smaller gain. On MENLO pairwise comparison and METAL summary rating, Qwen is the better receiving model, but the gains over Jev alone are only 0.40 and 0.68 percentage points, respectively. Confidence-based routing also outperforms random routing for both receiving models on all three tasks. The advantage is largest on Global-MMLU-Lite, where routing to GPT achieves 92.73\%, compared with 90.22\% under random selection. These results show that confidence helps select samples for routing, although the gains over Jev alone can be small. We therefore recommend using a development set to select the confidence threshold and the model that receives the routed samples when the budget allows, then applying the selected configuration to new samples. Under a limited budget, we recommend using Jev alone, given the small gains on most tasks.

\section{Conclusion}
\label{sec:conclusion}

We systematically evaluated Jev as a multilingual judge across four benchmarks. Jev achieves competitive performance on most tasks at substantially lower evaluation cost than GPT-6.1, and its probability outputs help identify incorrect judgments. Building on our analysis of its capabilities and limitations, we further evaluated practical configurations and developed recommendations on evaluation instructions, confidence-based selection, and selective routing.

\bibliography{references}

\appendix
\renewcommand{\topfraction}{0.95}
\renewcommand{\dbltopfraction}{0.95}
\renewcommand{\textfraction}{0.05}
\renewcommand{\floatpagefraction}{0.85}
\renewcommand{\dblfloatpagefraction}{0.85}
\setcounter{dbltopnumber}{4}
\setcounter{topnumber}{4}
\setcounter{totalnumber}{6}
\makeatletter
\setlength{\@fptop}{0pt}
\setlength{\@fpsep}{16pt}
\setlength{\@fpbot}{0pt plus 1fil}
\setlength{\@dblfptop}{0pt}
\setlength{\@dblfpsep}{16pt}
\setlength{\@dblfpbot}{0pt plus 1fil}
\makeatother
\section{Prompts}
\label{app:prompts}

\begin{figure*}[tp]
\centering
\includegraphics[width=\textwidth]{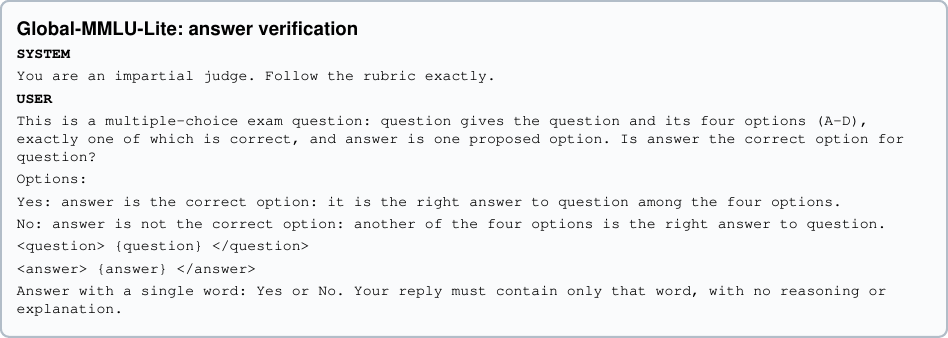}
\par\smallskip
\includegraphics[width=\textwidth]{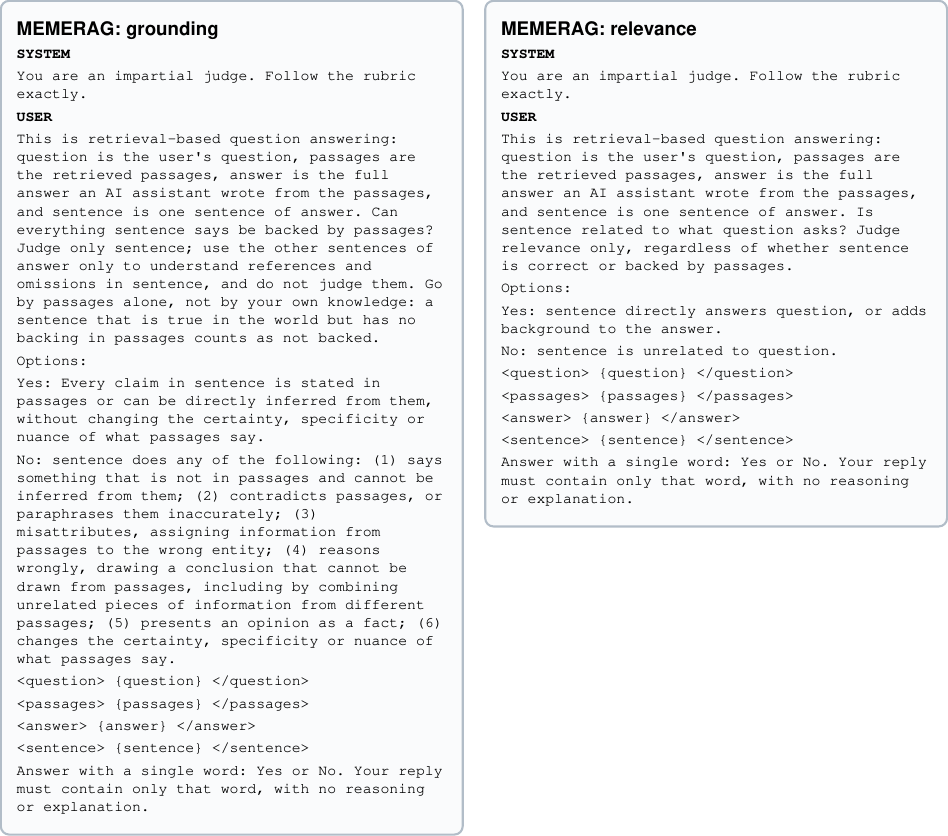}
\caption{Prompts for Global-MMLU-Lite answer verification (top) and MEMERAG sentence-level faithfulness evaluation (bottom).}
\label{fig:prompt-binary}
\end{figure*}

\begin{figure*}[tp]
\centering
\includegraphics[width=\textwidth]{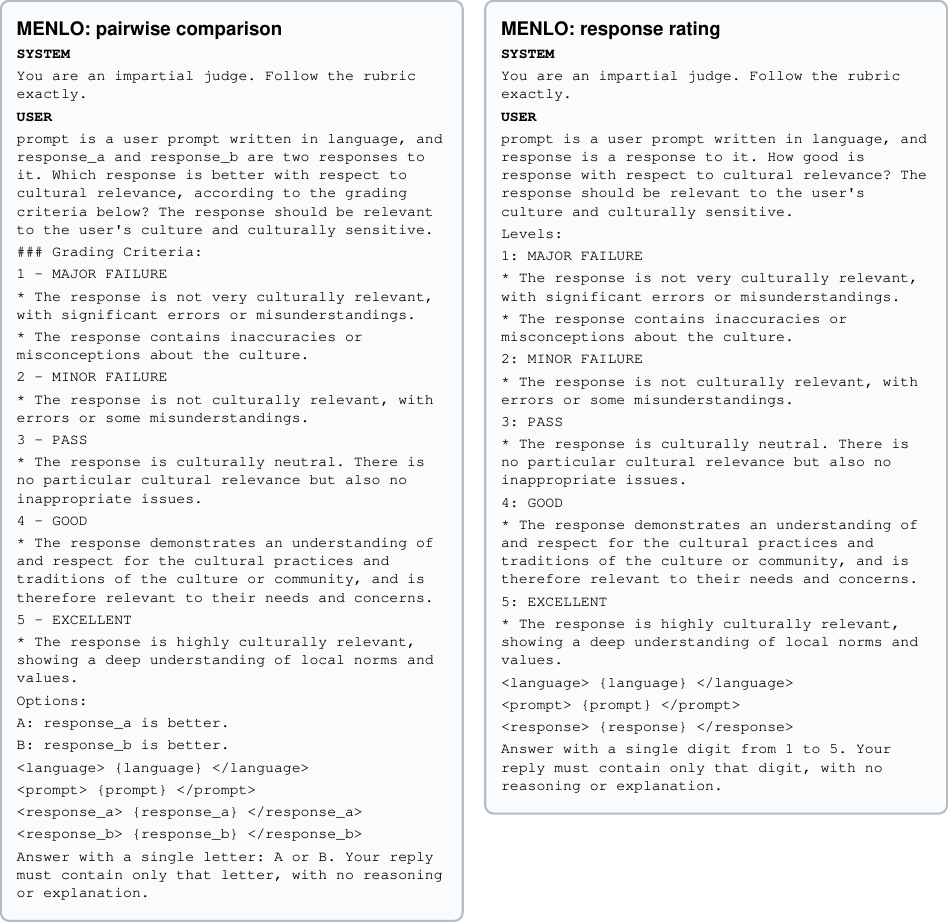}
\caption{MENLO prompts for pairwise comparison and response rating, shown for cultural relevance.}
\label{fig:prompt-menlo}
\end{figure*}

\begin{figure*}[tp]
\centering
\includegraphics[width=\textwidth]{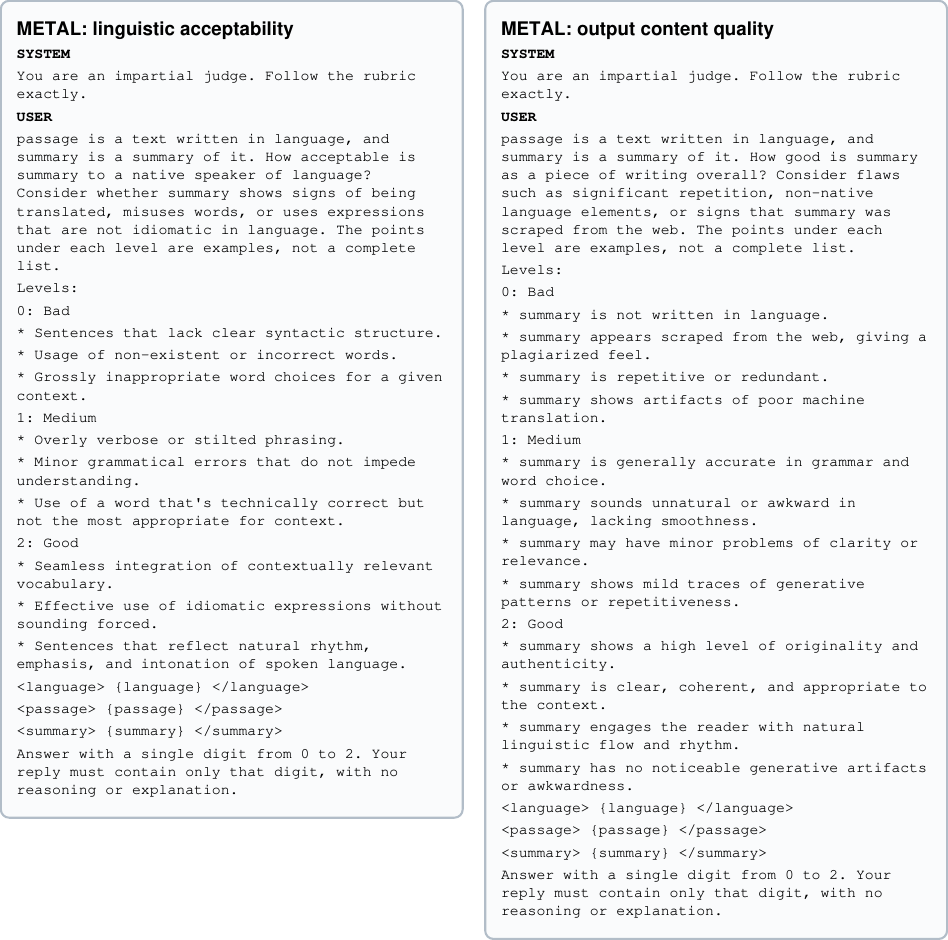}
\caption{METAL summary-rating prompts for linguistic acceptability and output content quality.}
\label{fig:prompt-metal-ratings}
\end{figure*}

\begin{figure*}[tp]
\centering
\includegraphics[width=\textwidth]{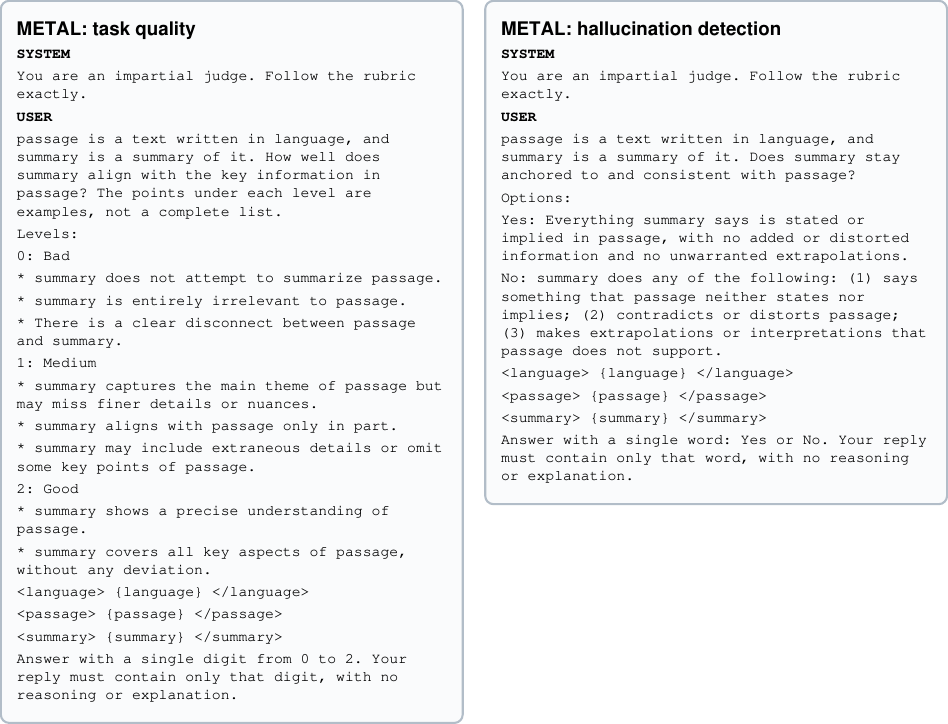}
\caption{METAL prompts for task-quality rating and hallucination detection.}
\label{fig:prompt-metal-task}
\end{figure*}

Figure~\ref{fig:prompt-binary} presents the prompts for Global-MMLU-Lite and MEMERAG, Figure~\ref{fig:prompt-menlo} for MENLO, and Figures~\ref{fig:prompt-metal-ratings}--\ref{fig:prompt-metal-task} for METAL.

\section{Data Processing}
\label{app:data}

\paragraph{MEMERAG.}
We evaluate individual sentences with the full response and retrieved passages as context. Support probability is the product of grounding and relevance probabilities. Jev evaluates both in one request; generative judges use separate requests. Sentences with indeterminate reference labels are excluded, leaving 2,034 test sentences shared across judges. We reserve 50 labeled sentences per language for development.

\paragraph{Global-MMLU-Lite.}
Each question yields two independent judgments, one for the correct answer and one for a randomly selected distractor shared across languages. We exclude questions with inconsistent answer labels, missing or corrupted content, duplicate options, or unavailable model outputs, applying exclusions across all languages and judges. The aligned test set contains 311 questions in 23 languages, separate from 47 development questions.

\paragraph{MENLO.}
Reference preferences are derived from individual human ratings; equal-rated pairs are excluded from pairwise evaluation. We reserve two unequal-rated pairs per language variety for development and exclude one pair exceeding a judge's context limit. The evaluation contains 5,166 preference pairs, each judged in both orders, and 12,658 individually rated responses across 47 varieties. We adapt the dataset instructions to each judgment format and remove cross-level references from Jev's rating criteria. Localized-factuality rubrics missing the lowest rating level are completed using the dataset's corresponding five-level rubric.

\paragraph{METAL.}
Summaries are matched to their source articles. Reference labels use the majority of three annotators, or the middle rating when all three differ. We reserve two summaries per language for development and exclude summaries with unavailable model outputs for all judges and dimensions, leaving 968 test summaries.

\section{Inference Details}
\label{app:inference}

Jev receives task instructions, evaluation criteria and inputs through its structured API and returns decision probabilities. Generative judges receive the corresponding text prompts and output discrete judgments, with thinking disabled for Qwen and low reasoning effort for GPT-6.1.

Qwen3.8-27B uses temperature 0; answer probabilities are obtained by renormalizing the top 20 first-token probabilities over valid answers. GPT-6.1 allows up to 8,000 output tokens, including internal reasoning; we use its discrete judgments without token probabilities.

\section{Per-Language Results}
\label{app:per-language}

Tables~\ref{tab:gmmlu-languages}, \ref{tab:memerag-languages}, \ref{tab:menlo-varieties}, \ref{tab:metal-languages} and~\ref{tab:metal-rating-languages} give every judge's results in each language of Global-MMLU-Lite, MEMERAG, MENLO and METAL.

\begin{table}[tbp]
\centering
\small
\setlength{\tabcolsep}{3pt}
\makebox[\columnwidth][c]{\begingroup\setbox0=\hbox{\begin{tabular}{lccccc}
\toprule
 & \multicolumn{2}{c}{Jev} & \multicolumn{2}{c}{Qwen} & GPT-6.1 \\
\cmidrule(lr){2-3} \cmidrule(lr){4-5} \cmidrule(lr){6-6}
Language & BAcc & AUROC & BAcc & AUROC & BAcc \\
\midrule
\multicolumn{6}{l}{\textit{High resource}} \\
Arabic & 90.7 & 95.7 & 85.0 & 92.5 & 94.7 \\
Chinese & 91.5 & 97.0 & 85.7 & 93.9 & 96.8 \\
English & 93.6 & 97.8 & 89.9 & 96.8 & 97.3 \\
French & 92.6 & 97.2 & 87.5 & 95.2 & 96.1 \\
German & 92.8 & 97.2 & 87.8 & 95.9 & 96.1 \\
Hindi & 88.6 & 95.6 & 77.8 & 86.6 & 96.1 \\
Italian & 92.8 & 97.4 & 88.7 & 95.1 & 96.9 \\
Japanese & 89.9 & 96.1 & 85.2 & 93.7 & 96.6 \\
Portuguese & 93.1 & 97.5 & 86.7 & 94.9 & 96.3 \\
Spanish & 93.9 & 97.5 & 87.3 & 95.2 & 96.6 \\
\midrule
\multicolumn{6}{l}{\textit{Mid resource}} \\
Bengali & 88.7 & 95.6 & 77.3 & 87.1 & 95.7 \\
Czech & 93.1 & 97.0 & 86.7 & 94.2 & 96.3 \\
Hungarian & 91.8 & 96.9 & 88.3 & 94.6 & 96.3 \\
Indonesian & 92.3 & 97.3 & 86.0 & 94.3 & 96.8 \\
Korean & 90.7 & 96.4 & 84.7 & 92.6 & 96.1 \\
Slovak & 93.2 & 97.5 & 88.7 & 95.2 & 96.6 \\
\midrule
\multicolumn{6}{l}{\textit{Low resource}} \\
Albanian & 92.9 & 97.4 & 87.1 & 94.9 & 96.6 \\
Burmese & 83.8 & 92.7 & 66.4 & 74.8 & 94.9 \\
Odia & 81.4 & 91.6 & 72.7 & 82.2 & 95.7 \\
Swahili & 83.1 & 93.6 & 76.2 & 84.6 & 95.2 \\
Tajik & 87.1 & 96.0 & 80.5 & 89.9 & 96.6 \\
Welsh & 87.8 & 95.9 & 80.9 & 90.4 & 96.8 \\
Yoruba & 74.9 & 86.6 & 68.8 & 78.4 & 93.1 \\
\bottomrule
\end{tabular}
}\ifdim\wd0>\columnwidth\resizebox{\columnwidth}{!}{\box0}\else\box0\fi\endgroup}
\caption{Answer-verification performance by language on Global-MMLU-Lite. Resource groups follow the dataset categories.}
\label{tab:gmmlu-languages}
\end{table}

\begin{table}[tbp]
\centering
\small
\setlength{\tabcolsep}{3pt}
\makebox[\columnwidth][c]{\begingroup\setbox0=\hbox{\begin{tabular}{lccccc}
\toprule
 & \multicolumn{2}{c}{Jev} & \multicolumn{2}{c}{Qwen} & GPT-6.1 \\
\cmidrule(lr){2-3} \cmidrule(lr){4-5} \cmidrule(lr){6-6}
Language & BAcc & AUROC & BAcc & AUROC & BAcc \\
\midrule
English & 67.3 & 78.4 & 70.8 & 78.1 & 72.1 \\
German & 66.7 & 81.4 & 72.5 & 81.3 & 73.1 \\
Spanish & 68.9 & 78.2 & 76.6 & 83.3 & 73.3 \\
French & 70.3 & 83.2 & 76.0 & 84.7 & 73.9 \\
Hindi & 71.9 & 83.9 & 74.6 & 82.8 & 75.1 \\
\bottomrule
\end{tabular}
}\ifdim\wd0>\columnwidth\resizebox{\columnwidth}{!}{\box0}\else\box0\fi\endgroup}
\caption{Sentence-level faithfulness evaluation by language on MEMERAG.}
\label{tab:memerag-languages}
\end{table}

\begin{table}[tbp]
\centering
\small
\setlength{\tabcolsep}{3pt}
\makebox[\columnwidth][c]{\begingroup\setbox0=\hbox{\begin{tabular}{lcccccc}
\toprule
 & \multicolumn{2}{c}{Jev} & \multicolumn{2}{c}{Qwen} & \multicolumn{2}{c}{GPT-6.1} \\
\cmidrule(lr){2-3} \cmidrule(lr){4-5} \cmidrule(lr){6-7}
Variety & Acc. & $\rho$ & Acc. & $\rho$ & Acc. & $\rho$ \\
\midrule
ar & 74.2 & $0.09$ & 71.3 & $0.10$ & 73.0 & $0.12$ \\
ar\_Latn\_EG & 85.7 & $-0.10$ & 91.4 & $-0.14$ & 92.9 & $-0.10$ \\
bg\_BG & 70.0 & $0.31$ & 67.8 & $0.37$ & 72.2 & $0.38$ \\
bn\_BD & 63.0 & $0.20$ & 63.0 & $0.25$ & 69.0 & $0.18$ \\
cs\_CZ & 75.4 & $0.28$ & 73.9 & $0.21$ & 75.7 & $0.27$ \\
da\_DK & 77.6 & $0.23$ & 73.0 & $0.27$ & 77.6 & $0.29$ \\
de\_DE & 56.4 & $0.41$ & 58.3 & $0.34$ & 60.3 & $0.36$ \\
el\_GR & 75.2 & $0.42$ & 76.0 & $0.46$ & 76.0 & $0.46$ \\
en\_AU & 75.0 & $0.21$ & 76.5 & $0.21$ & 72.8 & $0.26$ \\
en\_GB & 75.0 & $0.46$ & 75.0 & $0.42$ & 65.9 & $0.36$ \\
en\_IN & 80.2 & $0.42$ & 70.8 & $0.44$ & 65.6 & $0.38$ \\
en\_US & 66.3 & $0.46$ & 68.5 & $0.44$ & 69.6 & $0.38$ \\
es\_ES & 76.8 & $0.35$ & 74.4 & $0.34$ & 70.1 & $0.34$ \\
es\_MX & 79.0 & $0.40$ & 84.0 & $0.39$ & 76.5 & $0.37$ \\
fa\_IR & 80.6 & $0.34$ & 75.7 & $0.35$ & 79.3 & $0.35$ \\
fr\_FR & 75.5 & $0.17$ & 76.0 & $0.27$ & 65.6 & $0.13$ \\
gu\_IN & 73.0 & $0.20$ & 74.3 & $0.17$ & 64.3 & $0.16$ \\
he\_IL & 75.4 & $0.27$ & 80.3 & $0.38$ & 74.2 & $0.39$ \\
hi\_IN & 68.0 & $0.18$ & 65.4 & $0.22$ & 61.8 & $0.22$ \\
hi\_Latn\_IN & 81.6 & $0.27$ & 83.3 & $0.23$ & 70.5 & $0.19$ \\
hr\_HR & 69.2 & $0.12$ & 71.2 & $0.21$ & 69.2 & $0.24$ \\
hu\_HU & 62.8 & $0.10$ & 63.5 & $0.15$ & 61.3 & $0.10$ \\
id\_ID & 78.1 & $0.38$ & 77.7 & $0.37$ & 73.0 & $0.37$ \\
it\_IT & 70.4 & $0.15$ & 67.4 & $0.21$ & 63.9 & $0.09$ \\
ja\_JP & 72.2 & $0.36$ & 72.6 & $0.41$ & 70.4 & $0.40$ \\
ko\_KR & 81.6 & $0.50$ & 81.2 & $0.52$ & 82.4 & $0.49$ \\
mr\_IN & 76.1 & $0.32$ & 80.3 & $0.35$ & 71.0 & $0.32$ \\
ms\_MY & 77.0 & $0.39$ & 77.4 & $0.39$ & 73.0 & $0.36$ \\
ne\_NP & 73.6 & $0.10$ & 73.6 & $0.17$ & 75.3 & $0.13$ \\
nl\_NL & 72.0 & $0.35$ & 73.3 & $0.33$ & 71.1 & $0.26$ \\
pl\_PL & 71.1 & $0.08$ & 74.6 & $0.10$ & 70.7 & $0.13$ \\
pt\_BR & 82.4 & $0.40$ & 84.1 & $0.48$ & 75.3 & $0.40$ \\
pt\_PT & 82.1 & $0.57$ & 79.5 & $0.59$ & 75.0 & $0.47$ \\
ro\_RO & 73.8 & $0.25$ & 73.0 & $0.31$ & 75.0 & $0.29$ \\
ru\_RU & 55.1 & $0.17$ & 60.7 & $0.20$ & 63.6 & $0.20$ \\
sk\_SK & 78.0 & $0.32$ & 76.8 & $0.42$ & 79.3 & $0.42$ \\
sv\_SE & 68.1 & $0.16$ & 70.5 & $0.19$ & 55.7 & $0.10$ \\
sw\_KE & 76.8 & $0.15$ & 76.8 & $0.17$ & 75.2 & $0.16$ \\
th\_TH & 76.0 & $0.36$ & 75.5 & $0.13$ & 68.1 & $0.29$ \\
tl\_PH & 86.6 & $0.41$ & 83.9 & $0.45$ & 78.7 & $0.45$ \\
tr\_TR & 86.1 & $0.47$ & 83.6 & $0.48$ & 78.5 & $0.48$ \\
uk\_UA & 71.6 & $0.25$ & 72.4 & $0.33$ & 72.0 & $0.37$ \\
ur\_Latn\_PK & 78.0 & $0.28$ & 76.4 & $0.38$ & 75.2 & $0.30$ \\
ur\_PK & 77.0 & $0.23$ & 76.6 & $0.29$ & 78.6 & $0.33$ \\
vi\_VN & 74.2 & $0.27$ & 74.2 & $0.28$ & 76.6 & $0.30$ \\
zh\_CN & 73.9 & $0.36$ & 73.9 & $0.42$ & 69.8 & $0.34$ \\
zh\_TW & 80.0 & $0.49$ & 81.9 & $0.50$ & 73.7 & $0.42$ \\
\bottomrule
\end{tabular}
}\ifdim\wd0>\columnwidth\resizebox{\columnwidth}{!}{\box0}\else\box0\fi\endgroup}
\caption{MENLO performance by language variety. Acc.\ measures pairwise preference accuracy, and $\rho$ measures Spearman correlation between predicted and human response ratings.}
\label{tab:menlo-varieties}
\end{table}

\begin{table}[tbp]
\centering
\small
\setlength{\tabcolsep}{3pt}
\makebox[\columnwidth][c]{\begingroup\setbox0=\hbox{\begin{tabular}{@{}lccccc@{}}
\toprule
 & \multicolumn{2}{c}{Jev} & \multicolumn{2}{c}{Qwen} & GPT-6.1 \\
\cmidrule(lr){2-3}\cmidrule(lr){4-5}
Language & BAcc & AUROC & BAcc & AUROC & BAcc \\
\midrule
English & 79.6 & 91.6 & 75.2 & 92.0 & 76.9 \\
French & 90.8 & 97.3 & 87.0 & 98.0 & 81.0 \\
Chinese & 89.3 & 98.2 & 90.8 & 97.4 & 91.2 \\
Hindi & 77.7 & 87.5 & 74.8 & 88.0 & 71.7 \\
Arabic & 84.6 & 92.3 & 84.4 & 93.8 & 78.6 \\
Bengali & 79.1 & 75.7 & 77.5 & 70.6 & 72.5 \\
Russian & 89.4 & 93.6 & 88.9 & 94.2 & 75.3 \\
Turkish & 89.8 & 96.1 & 85.9 & 94.9 & 79.7 \\
Japanese & 90.7 & 97.9 & 94.4 & 98.4 & 84.0 \\
Swahili & 87.4 & 95.4 & 90.2 & 95.9 & 81.2 \\
\bottomrule
\end{tabular}
}\ifdim\wd0>\columnwidth\resizebox{\columnwidth}{!}{\box0}\else\box0\fi\endgroup}
\caption{METAL hallucination detection by language (\%).}
\label{tab:metal-languages}
\end{table}

\begin{table}[tbp]
\centering
\small
\begin{tabular*}{\columnwidth}{@{\extracolsep{\fill}}lccc@{}}
\toprule
Language & Jev & Qwen & GPT-6.1 \\
\midrule
English & 55.3 & 61.1 & 64.9 \\
French & 46.5 & 50.8 & 50.6 \\
Chinese & 48.4 & 45.9 & 48.1 \\
Hindi & 35.7 & 38.2 & 37.6 \\
Arabic & 49.3 & 48.8 & 45.3 \\
Bengali & 34.9 & 33.2 & 33.0 \\
Russian & 64.6 & 60.8 & 61.4 \\
Turkish & 54.0 & 57.5 & 55.9 \\
Japanese & 45.5 & 45.8 & 43.6 \\
Swahili & 61.5 & 72.3 & 58.0 \\
\bottomrule
\end{tabular*}

\caption{METAL summary rating by language. Scores are macro-F1 (\%) averaged across three evaluation dimensions.}
\label{tab:metal-rating-languages}
\end{table}

\section{Metric Aggregation}
\label{app:stats}

\paragraph{Binary verification.}
BAcc is the mean of positive- and negative-class recall within each language. BAcc and answer-label AUROC are then averaged equally across languages. Resource-group results use the same averaging within each group. For METAL hallucination detection, the positive label denotes a summary without hallucinations.

\paragraph{Pairwise comparison.}
We map predictions from both presentation orders to the same candidate and average their probabilities. A pair receives credit 1 for selecting the higher-rated response, 0 for the lower-rated response, and 0.5 for a tie. GPT-6.1's discrete choices are encoded as probabilities of 0 or 1. Accuracy is calculated within each language variety and then averaged equally across varieties. Order agreement compares the choices from the two presentation orders after mapping them to the same candidates, using probability-based choices for Jev and Qwen and discrete choices for GPT-6.1. Agreement is calculated within each language variety and averaged equally across the 47 varieties.

\paragraph{Graded rating.}
MENLO macro-F1 and Spearman correlation pool rated responses across languages and evaluation dimensions. For METAL, each metric pools languages within a dimension and is then averaged equally across the three dimensions. Results use discrete predicted ratings.

\section{Evaluation Cost}
\label{app:noise-cost}

Table~\ref{tab:noise-cost} reports total costs on the shared samples used for cost estimation. Costs are calculated from recorded token usage, including GPT-6.1's internal reasoning. The experimental prices per million tokens were \$0.042 for Jev input and output, and \$2 and \$10 for GPT-6.1 input and output, respectively. Qwen was run locally; its API-equivalent cost is estimated using Alibaba pricing. Table~\ref{tab:efficiency} reports the corresponding costs per 1,000 samples. Totals and per-sample costs are rounded independently.

\begin{table}[tbp]
\centering
\small
\setlength{\tabcolsep}{3pt}
\makebox[\columnwidth][c]{\begingroup\setbox0=\hbox{\begin{tabular}{@{}lrccc@{}}
\toprule
Benchmark & Samples & Jev & Qwen & GPT-6.1 \\
\midrule
MEMERAG & 2,035 & 0.17 & 2.53 & 10.20 \\
Global-MMLU-Lite & 14,553 & 0.38 & 2.33 & 11.88 \\
MENLO & 22,990 & 1.85 & 14.36 & 67.72 \\
METAL & 3,906 & 0.25 & 1.74 & 7.65 \\
\midrule
Total & 43,484 & 2.64 & 20.96 & 97.45 \\
\bottomrule
\end{tabular}
}\ifdim\wd0>\columnwidth\resizebox{\columnwidth}{!}{\box0}\else\box0\fi\endgroup}
\caption{Total evaluation costs (USD) on shared samples. Qwen costs use Alibaba API pricing.}
\label{tab:noise-cost}
\end{table}

\section{Analysis Details}
\label{app:analysis-details}
\subsection{Cross-lingual judgment transitions}
\label{app:cross-lingual-transitions}
Candidate judgments are matched to English by question ID and option. Each transition rate uses all candidates as its denominator, regardless of correctness in English. Rates are averaged equally across languages within each resource group, excluding English. The high-resource group contains ar, de, es, fr, hi, it, ja, pt and zh; the mid-resource group contains bn, cs, hu, id, ko and sk; the low-resource group contains cy, my, or, sq, sw, tg and yo. Overall rates weight these group means by 9, 6 and 7 languages, respectively.

The false-acceptance contribution to the Jev--GPT consistency gap is the difference in their false-acceptance rates divided by the difference in their total disagreement rates. Using unrounded values gives 78.3\% in the low-resource group.

\subsection{Confidence-based error identification}
\label{app:confidence-task-details}
For binary verification, confidence is $c=\max(p,1-p)$, where $p$ is the acceptance probability. MENLO preferences use the same formula after mapping both presentation orders to the original candidate A and averaging their probabilities. We exclude 19 pairs with $p=0.5$, leaving 5,147 pairs. For ratings, confidence is the probability of the selected level, and correctness requires an exact match with the reference rating.

Confidence AUROC pools all languages, including English, and all evaluation dimensions within each task. Each METAL summary--dimension pair counts as one judgment.

\section{Practical Guidelines: Experimental Details}
\label{app:guidance-details}

\subsection{Evaluation instruction language}
\label{app:native}
We compare instructions on matched samples: 13,684 Global-MMLU-Lite judgments across 22 non-English languages, and 976 MENLO preference pairs and 2,118 rated responses across eight languages and all ten dimensions. The MENLO languages are Arabic, German, Hindi, Hungarian, Japanese, Russian, Swahili and Chinese. Only task instructions and criteria are translated; evaluated content and labels remain unchanged.

Paired differences use 1,500 cluster-bootstrap resamples, grouping Global-MMLU-Lite by question across languages and options, and MENLO by prompt. Table~\ref{tab:app-instruction-language} reports paired differences and uncertainty estimates, including rating correlations. Input-token comparisons exclude translation costs.

\begin{table}[tbp]
\centering
\small
\setlength{\tabcolsep}{3pt}
\makebox[\columnwidth][c]{\begingroup\setbox0=\hbox{\begin{tabular}{@{}lcc@{}}
\toprule
Task  &  Metric  &  \shortstack{Target $-$\\English} \\
\midrule
\shortstack[l]{Global-\\MMLU-Lite}  &  BAcc  &  \shortstack{$-0.18$\\{}[$-0.39$, $0.04$]} \\
\shortstack[l]{MENLO\\preference}  &  Accuracy  &  \shortstack{$-0.39$\\{}[$-1.16$, $0.37$]} \\
\shortstack[l]{MENLO\\rating}  &  Macro-F1  &  \shortstack{$0.53$\\{}[$-0.65$, $1.72$]} \\
\shortstack[l]{MENLO\\rating}  &  Spearman $\rho$  &  \shortstack{$-0.005$\\{}[$-0.019$, $0.010$]} \\
\bottomrule
\end{tabular}
}\ifdim\wd0>\columnwidth\resizebox{\columnwidth}{!}{\box0}\else\box0\fi\endgroup}
\caption{Effect of translating evaluation instructions. Positive values favor target-language instructions over English. Brackets give 95\% cluster-bootstrap confidence intervals. Accuracy and F1 differences are in percentage points.}
\label{tab:app-instruction-language}
\end{table}

\subsection{Fixed confidence thresholds}
\label{app:confidence-selection}
Judgments with confidence $c\geq t$ are retained using the definitions in Appendix~\ref{app:confidence-task-details}, including the exclusion of MENLO probability ties. Accuracy and coverage pool all languages and evaluation dimensions.

\begin{table}[tbp]
\centering
\small
\setlength{\tabcolsep}{3pt}
\makebox[\columnwidth][c]{\begingroup\setbox0=\hbox{\begin{tabular}{@{}lcc@{}}
\toprule
Task & Threshold range & Routed (\%) \\
\midrule
Global-MMLU-Lite & 0.620 & 9.61 \\
MEMERAG & 0.579--0.591 & 10.08 \\
METAL (Hallucination) & 0.634--0.653 & 9.92 \\
MENLO (Preference) & 0.615--0.620 & 9.85 \\
MENLO (Rating) & 0.450 & 8.61 \\
METAL (Rating) & 0.550 & 9.33 \\
\bottomrule
\end{tabular}
}\ifdim\wd0>\columnwidth\resizebox{\columnwidth}{!}{\box0}\else\box0\fi\endgroup}
\caption{Confidence thresholds and actual test routing rates. The range covers thresholds selected from the lowest-confidence 10\% of development samples across five folds. ``Routed'' gives the percentage of test samples below those thresholds.}
\label{tab:app-routing-budgets}
\end{table}

\subsection{Development-set routing thresholds}
\label{app:routing-details}
We use five-fold content-disjoint evaluation, with four folds for development and one for testing. Groups are question IDs across languages and candidates for Global-MMLU-Lite, questions within language for MEMERAG, prompts for MENLO, and source articles for METAL.

The threshold is the development-set 10th percentile of confidence, applied unchanged to each test fold. Judgments below it are replaced by GPT or Qwen judgments in separate experiments. For the random-routing baseline, we sample without replacement once within each test fold, matching the exact number of low-confidence samples routed. We use seed 87123 and the same selected samples for both receiving models. Confidence ties are not split, so actual routing proportions differ slightly from 10\%. Results aggregate held-out predictions using the main evaluation metrics. Routing includes all 5,166 MENLO preference pairs, with half credit for tied decisions. Table~\ref{tab:app-routing-budgets} gives the thresholds and actual test routing proportions.

\end{document}